\documentclass{article}

\usepackage[preprint,nonatbib]{neurips_2023}

\usepackage[utf8]{inputenc} 
\usepackage[T1]{fontenc}    
\usepackage{url}            
\usepackage{booktabs}       
\usepackage{amsfonts}       
\usepackage{nicefrac}       
\usepackage{microtype}      
\usepackage{xcolor}         

\usepackage[outline]{contour}
\usepackage{amsmath,amssymb,amsfonts}
\usepackage{tikz}
\usetikzlibrary{shapes.geometric, arrows}
\usepackage{hyperref}       
\usepackage{cleveref}

\newcommand{\x}{\mathbf{x}}

\title{Inventing art styles with no artistic training data}

\author{
Nilin Abrahamsen\\
The Simons Institute for the Theory of Computing\\
Berkeley, CA, USA\\
\texttt{nilin@berkeley.edu}
\AND
Jiahao Yao\\
Department of Mathematics\\
University of California, Berkeley\\
Berkeley, CA, USA\\
\texttt{jiahaoyao@berkeley.edu}
}
\date{May 2023}

\begin{document}

\maketitle

\begin{abstract}
We propose two procedures to create painting styles using models trained only on natural images, providing objective proof that the model is not plagiarizing human art styles. In the first procedure we use the inductive bias from the artistic medium to achieve creative expression. Abstraction is achieved by using a reconstruction loss. The second procedure uses an additional natural image as inspiration to create a new style. These two procedures make it possible to invent new painting styles with no artistic training data. 
We believe that our approach can help pave the way for the ethical employment of generative AI in art, without infringing upon the originality of human creators.
\end{abstract}

\section{Introduction}
Recent advances in AI raise important questions about the essence of human creativity and the future trajectory of art work. 
In the field of visual arts, products such as Midjourney and Dall-E are generating images that arguably pass as human-made art with little effort from the user. These models have been trained on millions of images and artworks from the internet, and many are of the opinion that the models essentially plagiarize the art styles that they have consumed through their training process. Different approaches have been proposed in response to the concern of AI plagiarizing art, including:
\begin{enumerate}
\item\label{it:adv} \textbf{Cloaking with adversarial perturbations \cite{shan_glaze_2023,liang_adversarial_2023}.} Artists who wish to protect themselves from plagiarism by AI may attempt to perturb their artworks in a way that is imperceptible to the human viewer but is meant to foil the AI training.
\item \textbf{Combing through training data.} Since many artists do not consent to the use of their artworks to train AI, services have appeared which offer to search through datasets to expose use of an artist's work as training data \cite{edwards_have_2022}.
\item \textbf{Through copyright law.} A recent class-action lawsuit~\cite{woods_artists_2023,chayka_is_2023} sued Midjourney Inc, DeviantArt Inc, and Stability A.I. for using artists' work without their consent. 
\end{enumerate}

\begin{figure}[h]
  \centering
  \begin{tikzpicture}
\node (painting){
\includegraphics[width=.6\textwidth,height=.5\textwidth]{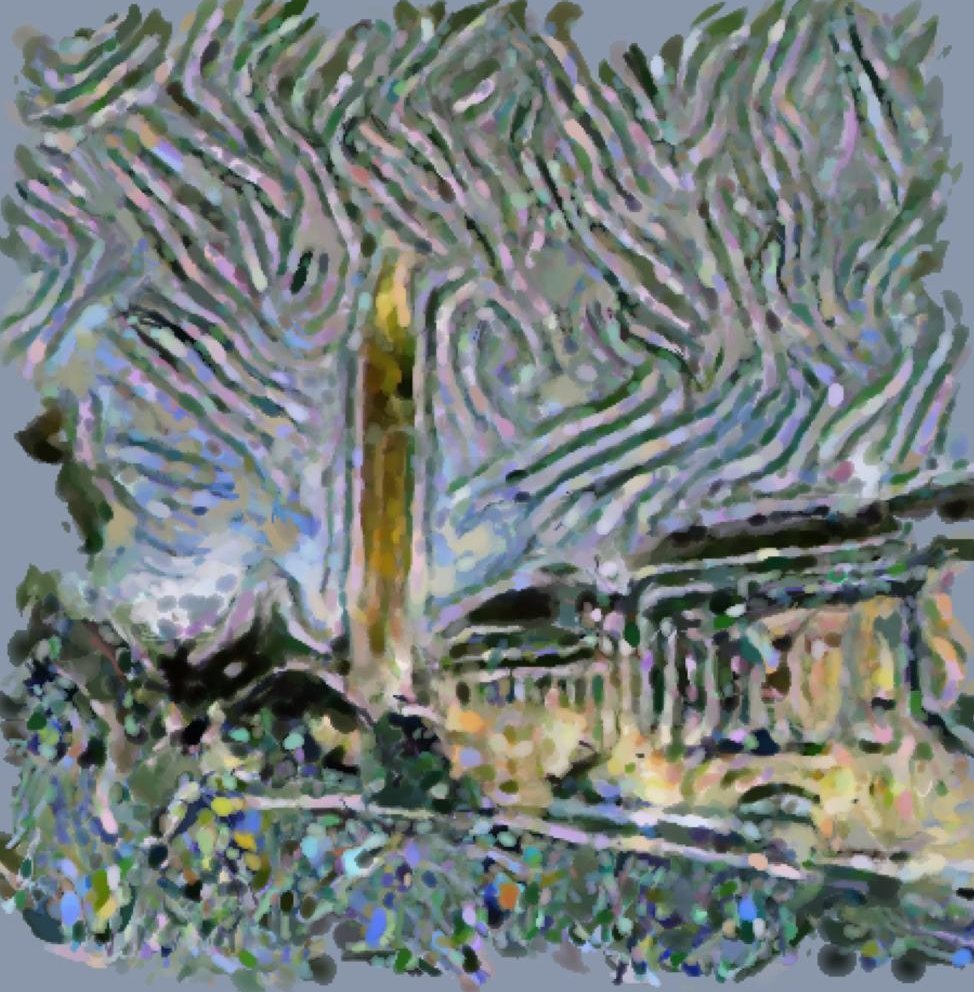}
};
\node (insp)[right of=painting,xshift=3.5cm,yshift=-3cm]{
\fcolorbox{black}{white}{\includegraphics[height=.12\textwidth,width=.14\textwidth]{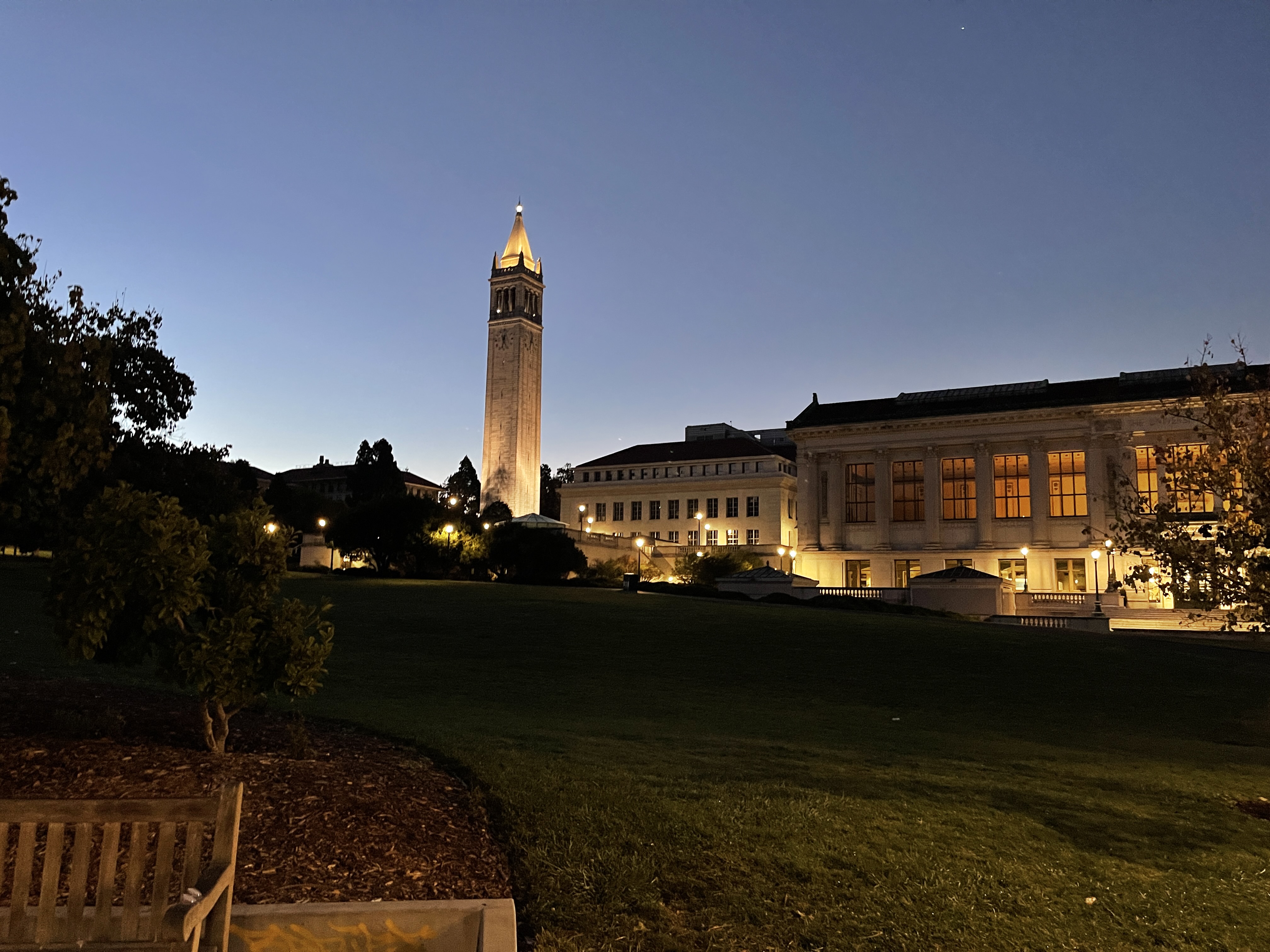}}
};
\node[below of=insp,yshift=-.18cm]{{\small Subject}};
\node (insp)[right of=painting,xshift=4.5cm,yshift=-0.8cm]{
\fcolorbox{black}{white}{\includegraphics[height=.2\textwidth]{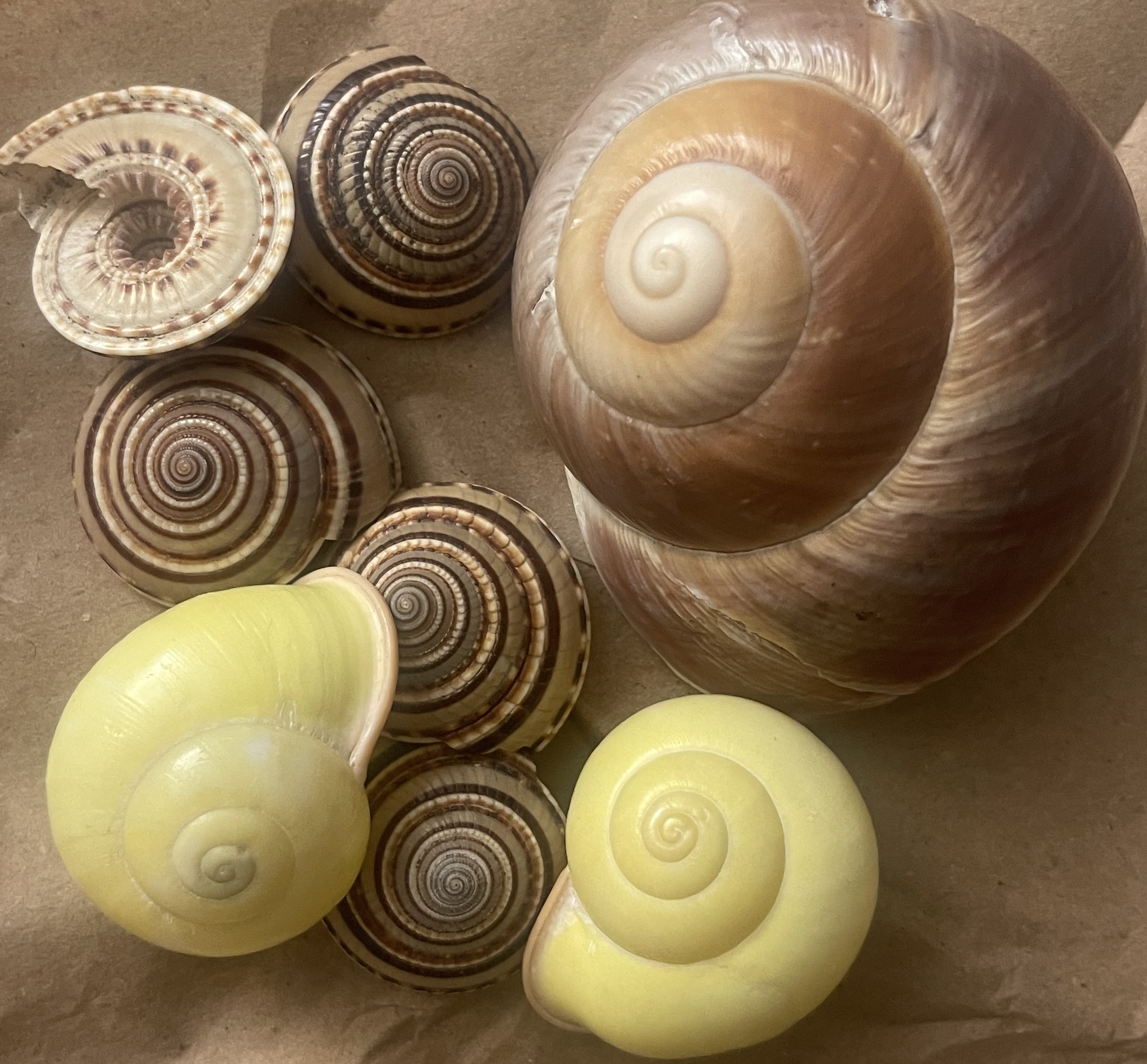}}
};
\node[above of=insp,yshift=.7cm]{
\textbf{\emph{Inspiration}}};
\end{tikzpicture}
  \caption{A tower painted with a combination of the two procedures proposed in this paper. Both inspiration and subject images are photographs by the authors.}\label{firstpic}
\end{figure}



The cloaking method is likely to be brittle in the long term as it relies on assumptions about how the AI processes its training data and how this differs from the intended human audience. 
Meanwhile the pushback and lawsuits against training by AI illustrate a long-standing discussion about the internal workings in a neural network. The plaintiffs' side of the previously mentioned lawsuit claims that \textit{``Stable Diffusion relies on a mathematical process called diffusion to store compressed copies of these training images, which in turn are recombined to derive other images. It is, in short, a 21st-century collage tool.''} \cite{butterick_stable_nodate} Although this notion that an AI model ``recombines'' the training data is generally not considered accurate by AI researchers and practitioners, it is however very difficult to rule out that this occurs at least for a small subset of the training data. This is especially true as modern models frequently have billions of parameters in which training data could hide. Indeed, \cite{carlini_extracting_2023} was able to extract training images from diffusion models such as Stable Diffusion using text prompts.



\subsection{Our contribution}
We propose two procedures to create painting styles using models trained only on natural images. This provides objective proof that the model does not plagiarize art styles made by humans.

The first procedure achieves creative expression through the \emph{inductive bias} from a chosen \emph{artistic medium}. We combine this with the flexibility of using a reconstruction loss to allow \emph{abstraction}. This first procedure can be viewed as a variant of image-to-image translation for a setting where we have no samples from the target domain. That is, the style itself is trainable and is generated by the artist through experimenting with the artistic medium. The preferred styles will be those which are able to be decoded to reconstruct the input image under the constraints of the artistic medium. We call this the medium+perception-driven procedure.

Our second procedure allows the algorithm to make use of natural images as \emph{inspiration} to create new painting styles. The use of inspiration from the natural world means that the creation of art styles can be guided by the user even though the model does is not exposed to human-made art. We call this the inspiration procedure.

Generative AI models are currently under attack for plagiarizing training data. Ironically, our proposal illustrates that they can in principle be used in a way to objectively avoid plagiarism by restricting their training data, something that would be infeasible for human creators. 
We include a discussion about possible implications at the end of the paper. 


\begin{figure}[h]
  \centering
  \begin{tikzpicture}
  
\node (painting){
\includegraphics[width=.4\textwidth,height=.4\textwidth]{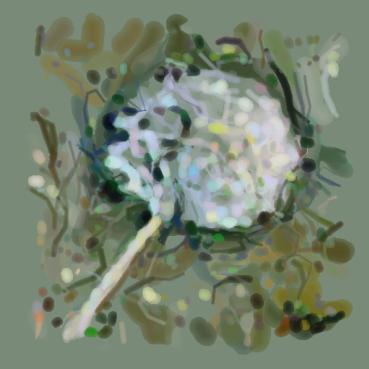}
};
\node (insp)[right of=painting,xshift=1.5cm,yshift=-2cm]{
\fcolorbox{black}{white}{\includegraphics[height=.15\textwidth]{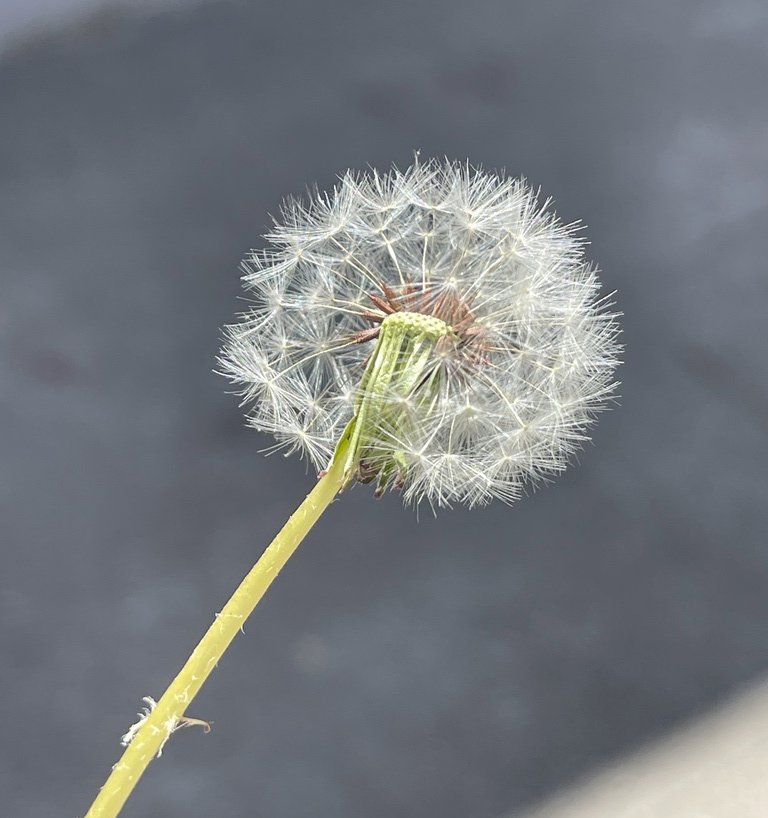}}
};

\node (painting2)[right of=painting,xshift=6cm]{
\includegraphics[width=.4\textwidth,height=.4\textwidth]{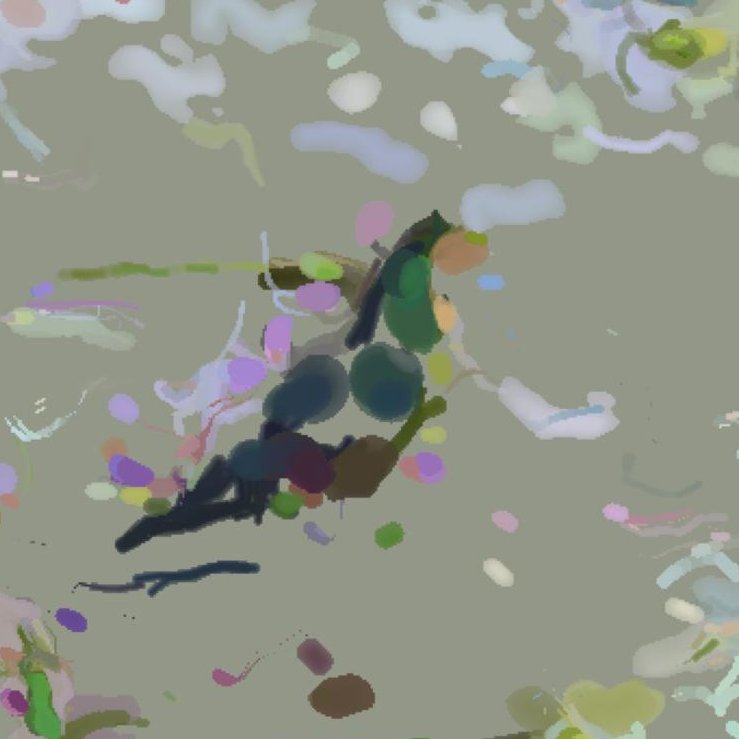}
};
\node (insp)[right of=painting2,xshift=1.5cm,yshift=-2cm]{
\fcolorbox{black}{white}{\includegraphics[height=.15\textwidth]{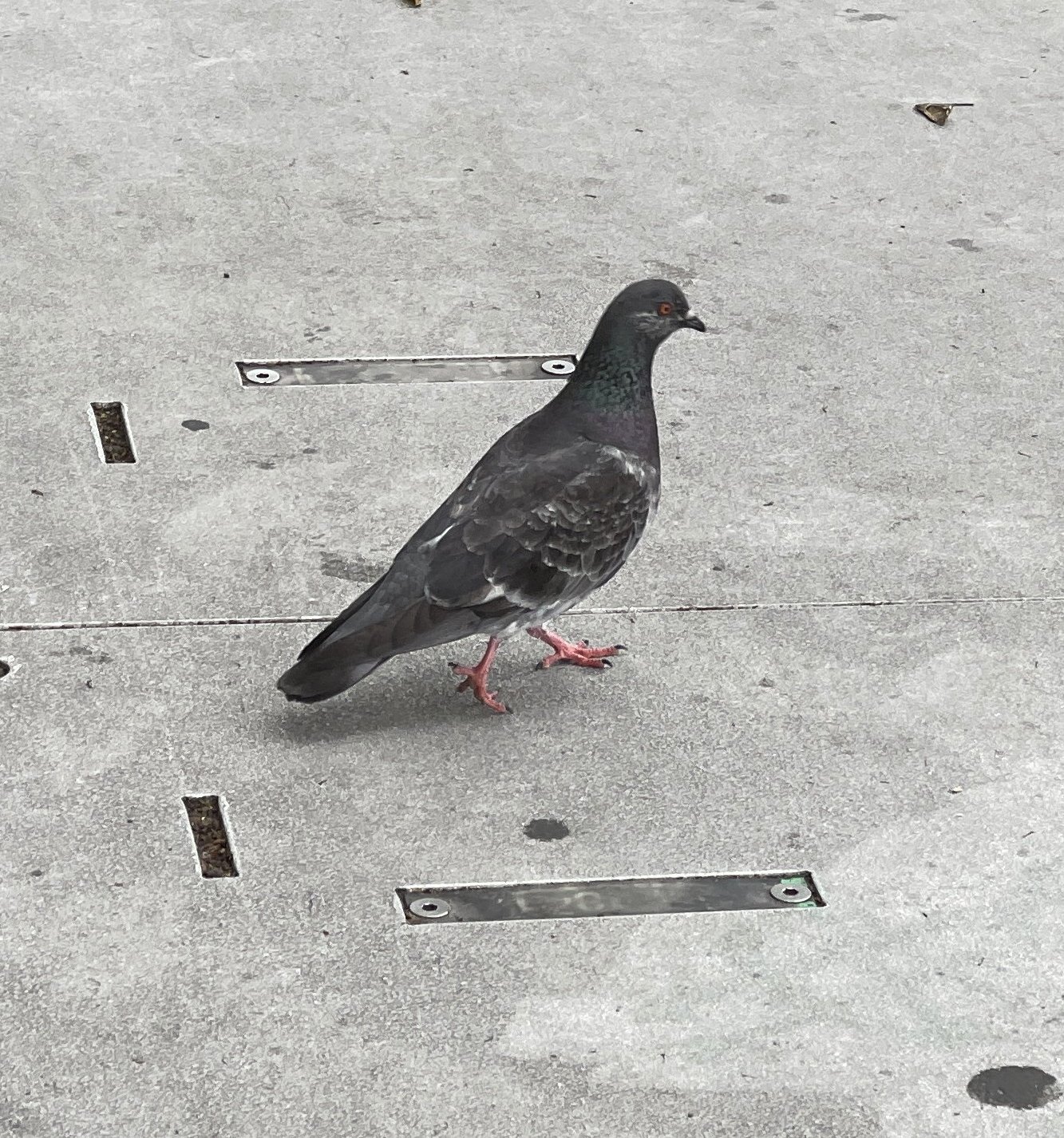}}
};

\end{tikzpicture}
  \caption{A dandelion and a pigeon painted using the medium+perception-driven procedure. Subject images are shown in the corner of each painting. All training images and subject images were photographs by the authors, ensuring that no artworks were present in the training data.}
  \label{pigeon}
\end{figure}

\subsection{Prior work}

\textbf{Algorithmic painting.} The concept of computer-generated artwork emerged as early as 1990, when  \cite{haeberli_paint_1990} implemented a brush engine and devised various ways of deciding the parameters (position, direction, color, etc.) of the brush strokes. This innovative approach included several methods: (1) interactively chosen brush strokes through user input, (2) randomly positioned brush strokes with color and direction based on the reference image. More advanced techniques included: (3) a painting with a 3D model as a reference where the direction of the brush strokes was based on the orientation of the 3D surface as determined by ray tracing and (4) which used iterative relaxation to approximate the subject image in L2-norm with rectangles or Dirichlet domains.

Our medium+perception-driven procedure can be viewed as an analogue of method (4) described earlier. We employ a reconstruction loss to facilitate \emph{abstraction} in the artwork while learning a mapping from subject images to paintings. The latter requires us to create an encoding of the artist's actions which can be produced as the output of a convolutional neural network.

\textbf{Style Transfer.}
The concept of style transfer was first advanced by Gatys et al.~\cite{gatys2016image}. Their method leverages convolutional neural networks to transfer the stylistic features of one image, referred to as the style source, onto another, known as the subject image. This technique effectively amalgamates the style and content from different images to create novel visual outputs. In addition, CycleGAN~\cite{zhu2017unpaired} is an \emph{unpaired} image-to-image translation model which generates a \emph{bijection} between two domains (or styles) $\mathcal X$ and $\mathcal Y$. It was revolutionary for its success in achieving this task using unpaired data, thereby eliminating the need of a one-to-one mapping between source and target domain images in the training set. CycleGAN learns two maps $G:\mathcal X\to\mathcal Y$ and $F:\mathcal Y\to\mathcal X$ and leverages \emph{cycle consistency losses} $d(F(G(x),x)$ and $d(G(F(y),y)$ to ensure that the maps are the inverses of each other. The method employs adversarial discriminators~\cite{goodfellow2014generative} on each of $\mathcal X$ and $\mathcal Y$ to ensure that the distribution of data $x\in\mathcal X$ and generated images $y\in\mathcal Y$ are matched within their appropriate domains.

A number of works have employed style transfer with more direct control of the geometry to control the rendering of pen and brush strokes \cite{reimann_controlling_2022,chen_anisotropic_2020,chan_learning_2022}.
The Stroke Control Multi-Artist Style Transfer framework~\cite{chen_anisotropic_2020} features an Anisotropic Stroke Module that allows for dynamic style-stroke adjustments. It also introduces a novel Multi-Scale Projection Discriminator for texture-level conditional generation. This enables the transformation of a photograph into various artistic style oil paintings, while preserving unique artistic style and anisotropic semantic information. Additionally, the work by Chan et al.~\cite{chan_learning_2022} proposes an unpaired method for generating line drawings from photographs. This process incorporates a geometry loss to predict depth information and a semantic loss to match features between a line drawing and its corresponding photograph.

Our inspiration-driven method is related to these works but differs in that it does not require examples of existing art styles. 


\textbf{Generative models.}
Generative modeling is a machine learning approach that aims to either generate new samples that are similar to the training data or learn the underlying probability density from the data. It is often categorized as a form of unsupervised or self-supervised learning. Prominent examples of generative models include Variational Autoencoders (VAEs)~\cite{kingma2013auto}, Generative Adversarial Networks (GANs)~\cite{goodfellow2014generative}, and Normalizing Flows \cite{rezende2015variational} and Vector-quantized Image Modeling (VIM) approach like VQGAN, VQVAE~\cite{van2017neural, yu2021vector}. 

\textbf{Diffusion models.}
Diffusion models are currently among the tools at the forefront of generative modelling.   
The pioneering work by Johnson Ho, et al. introduced some of the first diffusion models~\cite{NEURIPS2020_4c5bcfec}. Diffusion models alternate between injecting noise and projecting onto the space of valid samples. They are generally considered easy to train to get high-quality data. Variants of the diffusion are Denoising Diffusion Implicit Models
 (DDIM)~\cite{song2020denoising} and cascaded diffusion models~\cite{ho2022cascaded}.

\tikzstyle{wideblock} = [rectangle, minimum width=2.5cm, minimum height=.7cm,text centered, fill=blue!10,draw=blue!30]

\tikzstyle{wideblock1} = [rectangle, minimum width=2.5cm, minimum height=.7cm,text centered, fill=red!10,draw=blue!30]
\tikzstyle{wideblock2} = [rectangle, minimum width=2.5cm, minimum height=.7cm,text centered, fill=blue!10,draw=blue!30]

\tikzstyle{block} = [rectangle, rounded corners, minimum width=1.5cm, minimum height=.7cm,text centered, draw=black, fill=red!30]
\tikzstyle{arrow} = [thick,->,>=stealth]

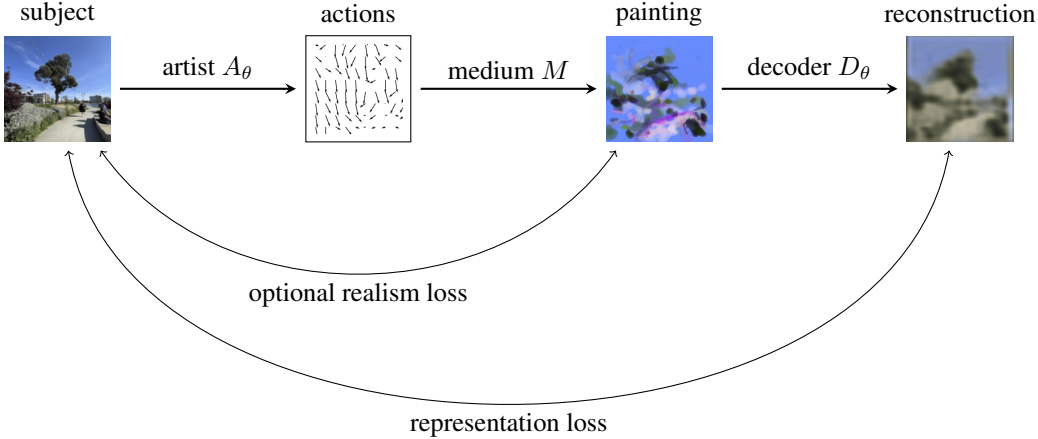
\begin{figure}[h]
  \centering
  \begin{tikzpicture}
\node (ref) [] {
\includegraphics[height=.1\textwidth]{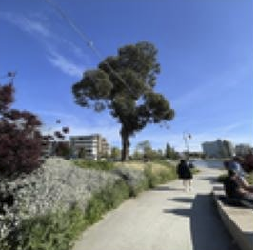}
};
\node (lat) [right of=ref, xshift=3cm] {
\fbox{\includegraphics[height=.085\textwidth]{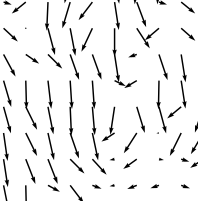}}
};
\node (pai) [right of=lat, xshift=3cm] {
\includegraphics[height=.1\textwidth]{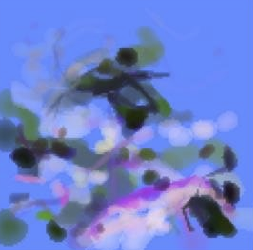}
};
\node (rec) [right of=pai, xshift=3cm] {
\includegraphics[height=.1\textwidth]{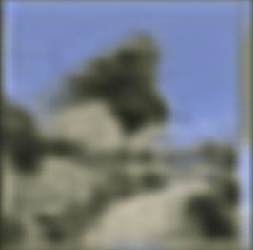}
};
\draw [arrow] (ref) --node[above]{artist $A_\theta$} (lat);
\draw [arrow] (lat) --node[above]{medium $M$} (pai);
\draw [arrow] (pai) --node[above]{decoder $D_\theta$} (rec);
\draw [<->] (ref) to [out=-80,in=-100]node[below]{representation loss} (rec);
\draw [<->] (ref) to [out=-55,in=-125]node[below]{optional realism loss} (pai);

\node[above of=ref,yshift=0cm]{
subject
};
\node[above of=lat,yshift=0cm]{
actions
};
\node[above of=pai,yshift=0cm]{
painting
};
\node[above of=rec,yshift=0cm]{
reconstruction
};
\end{tikzpicture}
  \caption{Creating artistic expression through abstract representation under the constraint of the artistic medium}
\end{figure}

\section{Medium+perception-driven procedure}

In our first approach, the creation of artistic styles is guided by the \emph{artistic medium}. We model the medium as a fixed function $M:\mathcal A\mapsto\mathcal P$ which maps a set of \emph{actions} $a\in \mathcal A$ (say, the coordinates of brush strokes) to a finished product $p\in\mathcal P$. Let $\mathcal S$ be the domain of subjects (natural images). To illustrate our ideas the focus on representational art using a paintbrush as the chosen medium. The elements of our first procedure are:

\begin{enumerate}
\item\label{it:medium} \textbf{Deliberate use of artistic medium.}
Staying with the paintbrush as an example, it would be possible to recreate a subject image to high accuracy by essentially \emph{printing} individual pixels of the image using small dabs of the paintbrush. However, this approach does not make efficient use of the brush and the shapes that it is able to make, resulting in a set of actions $a$ of high complexity. We propose that the process of optimizing a loss $\ell(p)=\ell(M(a))$ under the constraints imposed by the artistic medium is an element of creative expression which which we can simulate by bounding the number of latent variables $\operatorname{dim}(a)$.

\item \textbf{Interpretable abstraction.}
In order to ensure that the painting $p$ is an abstract representation of the subject $s$ we employ a \emph{reconstruction} loss defined as $\tilde\ell_\theta(p,s)=d(D_\theta(p),s)$ where $d$ is a distance function on the space of images and $D_\theta$ is a trainable decoder. We also add a tuneable $l_1$-loss directly between the painting and the subject and the painting where the parameter $\beta$ is tuned to adjust the realism of the painting.
\end{enumerate}


The decoder in principle models a bijection between the set of paintings $M\circ A_\theta(\mathcal S)$ and the set of subjects $\mathcal S$ \footnote{That is, it is a bijection in the reconstruction loss is $0$.}, and we think of this bijection as a simple proxy for the artist's \emph{perception} \cite{gombrich_art_2000}. Gombrich also argued for the appeal of simplicity in art \cite{gombrich_preference_2002}, motivating \cref{it:medium}.

We train a parameterized \emph{artist} $A_\theta$ to generate the actions $a$ given a subject image $s$. The full loss function is thus:
\[\ell_\theta(s)=d_{\operatorname{rec}}(D_\theta(M\circ A_\theta(s)),s)+\beta d_{\operatorname{realism}}(M\circ A_\theta(s),s).\]
For \cref{pigeon} we used
$d_{\operatorname{rec}}(A,B)=\log\|A-B\|_1+\log\operatorname{Dirdist}(A,B)$ and $d_{\operatorname{realism}}=\log\|A-B\|_1$, $\beta=1$.
where $\operatorname{Dirdist}$ is a distance measure that compares the local geometry of $A$ and $B$ and which we describe in detail in \cref{dirloss}. Note that $\operatorname{Dirdist}$ is only applied between the subject image and the reconstruction which should both exist in the space of natural images, so we are not directly guiding the style of the painting by adding this loss term.
    

\textbf{Relation to autoencoders and CycleGAN}

Our setting can be viewed as a version of this image translation problem where we have no samples from domain $\mathcal Y$. Instead we have a map $M$ (the artistic medium) whose outputs are in $\mathcal Y$. Put differently, the elements of $\mathcal Y$ are generated by the artist through the $M\circ A_\theta$ and are trainable. Thus, our representation loss is analogous to the cycle consistency loss $d(F(G(x)),x)$ in CycleGAN. We do not need an analogue of the style disctriminators because
\begin{itemize}
    \item Our outputs belong to the domain of paintings by design, and
    \item The painting style not fixed but a product of the training dynamics.
\end{itemize}
Optionally, we could train a discriminator to learn the divergence distance between the reconstructed images and the original ones. 
The realism loss has no analogue in CycleGAN, and we add it to compensate for the large flexibility from the lack of a target style. We find that it helps preserve the coloring of the images. For example, without it the artist would invert the brightness values with 50 percent chance.

Our procedure can also be viewed as a version of an \emph{auto-encoder} where the latent variables are interpretable either as the actions of the artist or as a painting. In the former case the decoder factors through the artistic medium $M$, and in the latter case the encoder factors through $M$.

\begin{figure}[h]
  \centering
  \begin{tikzpicture}
\node (painting){
\includegraphics[width=.3\textwidth,height=.3\textwidth]{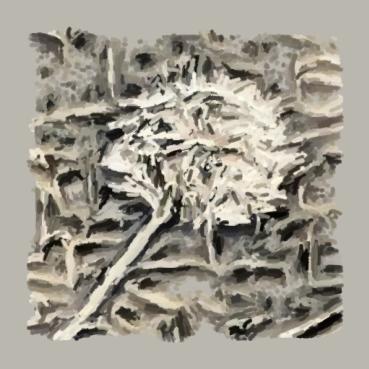}
};
\node (insp)[right of=painting,xshift=0cm,yshift=-2cm]{
\fcolorbox{black}{white}{\includegraphics[height=.1\textwidth]{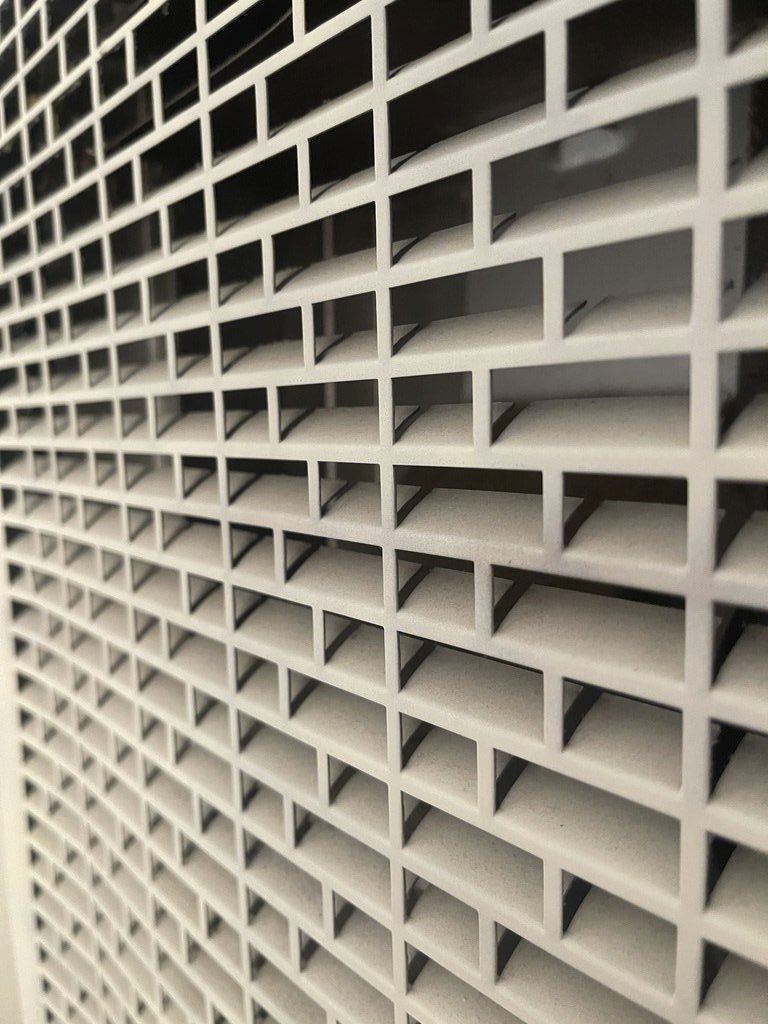}}
};
\end{tikzpicture}
  \begin{tikzpicture}
\node (painting){
\includegraphics[width=.3\textwidth,height=.3\textwidth]{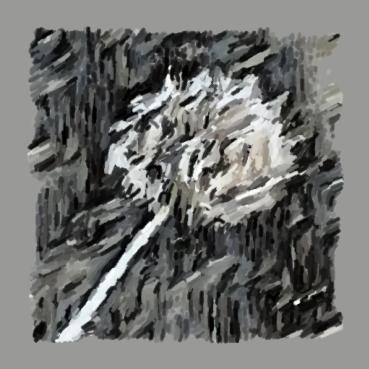}
};
\node (insp)[right of=painting,xshift=0cm,yshift=-2cm]{
\fcolorbox{black}{white}{\includegraphics[height=.1\textwidth]{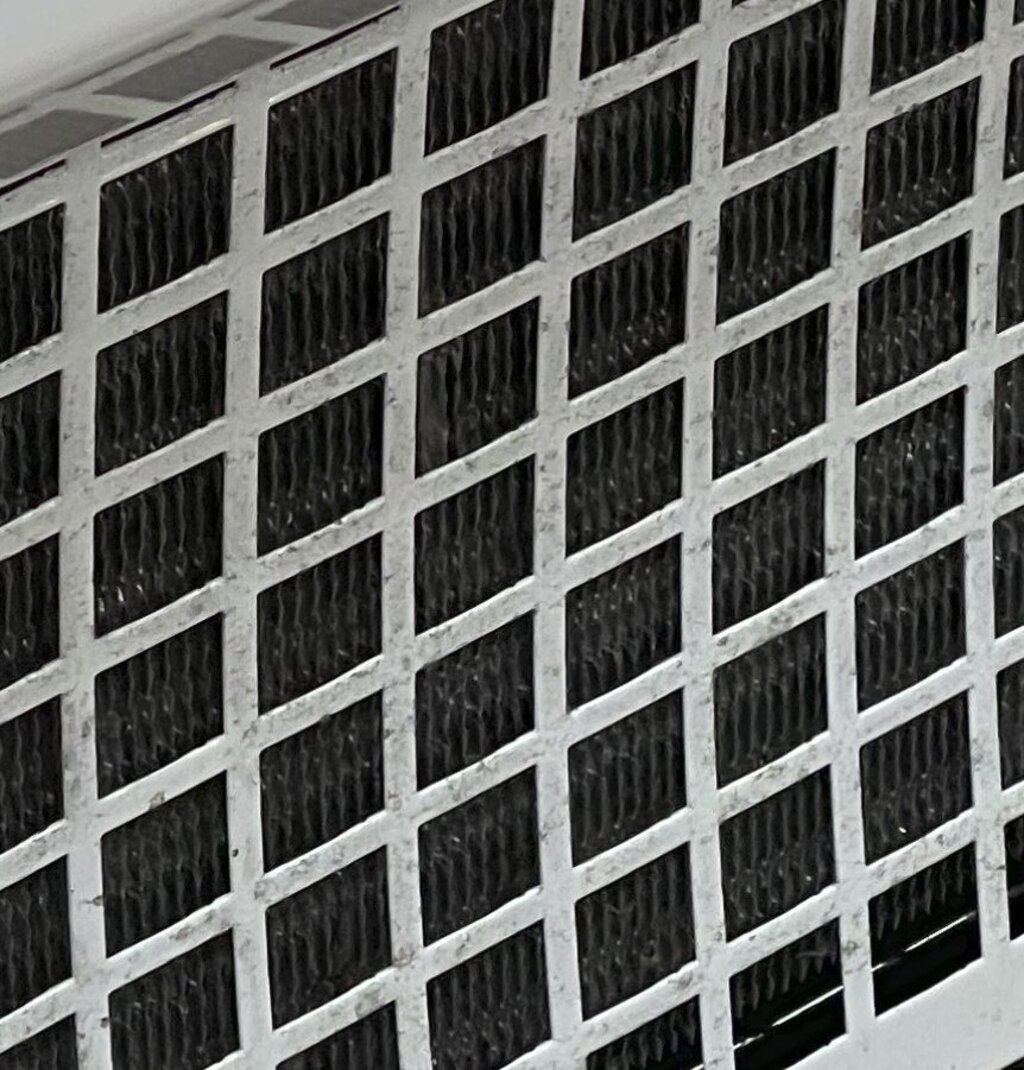}}
};
\end{tikzpicture}
  \begin{tikzpicture}
\node (painting){
\includegraphics[width=.3\textwidth,height=.3\textwidth]{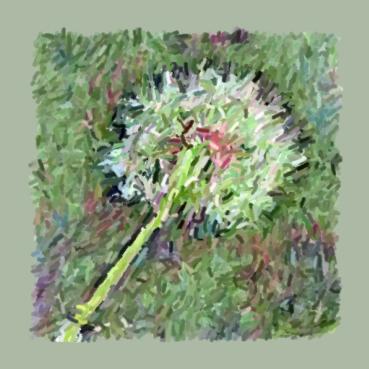}
};
\node (insp)[right of=painting,xshift=0cm,yshift=-2cm]{
\fcolorbox{black}{white}{\includegraphics[height=.1\textwidth]{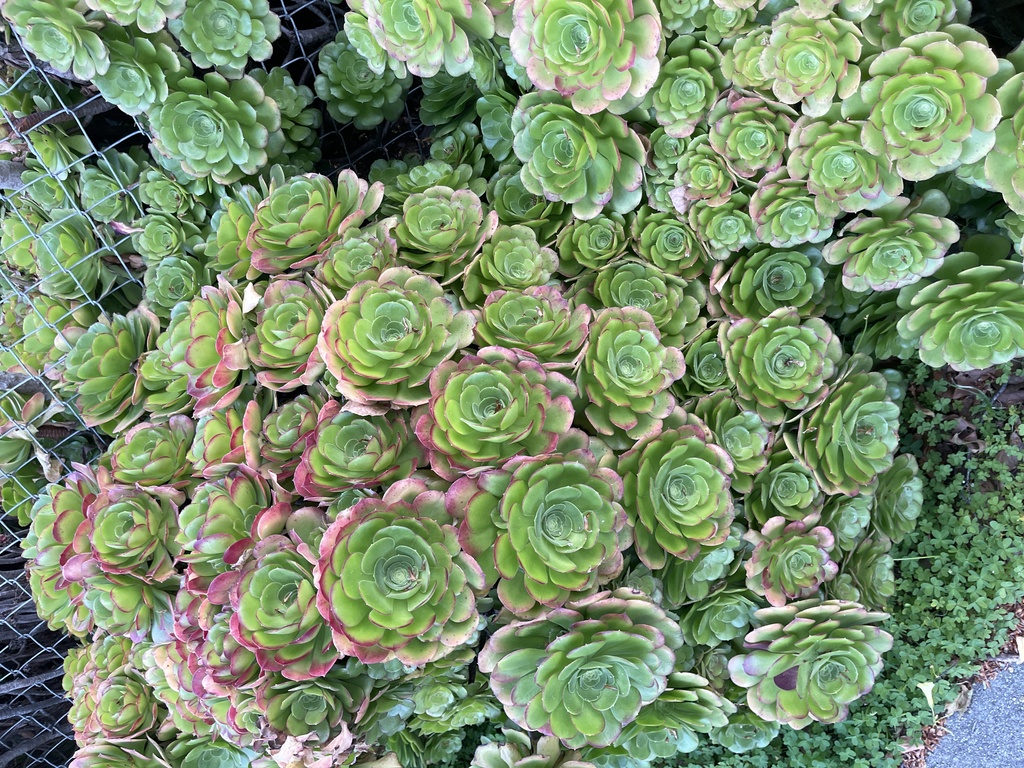}}
};
\end{tikzpicture}
  \caption{Different painting styles with the same baseline technique and different inspiration photographs (the bottom-right frame). To illustrate the inspiration procedure in an isolated manner we use a simple hard-coded baseline technique in \cref{inspcomp} and \cref{inspillustration}. }
  \label{inspcomp}
\end{figure}

\begin{figure}[h]
  \centering
  \begin{tikzpicture}
\node (mid){
\includegraphics[width=.1\textwidth,height=.1\textwidth]{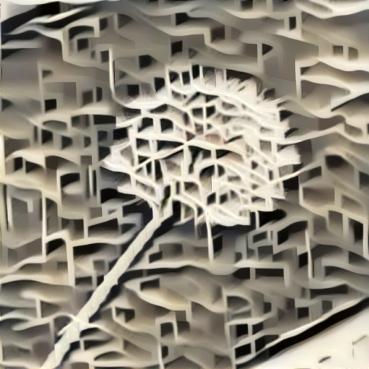}
};
\node (ins) [left of=mid, yshift=2cm,xshift=-2cm]{
\includegraphics[width=.2\textwidth,height=.2\textwidth]{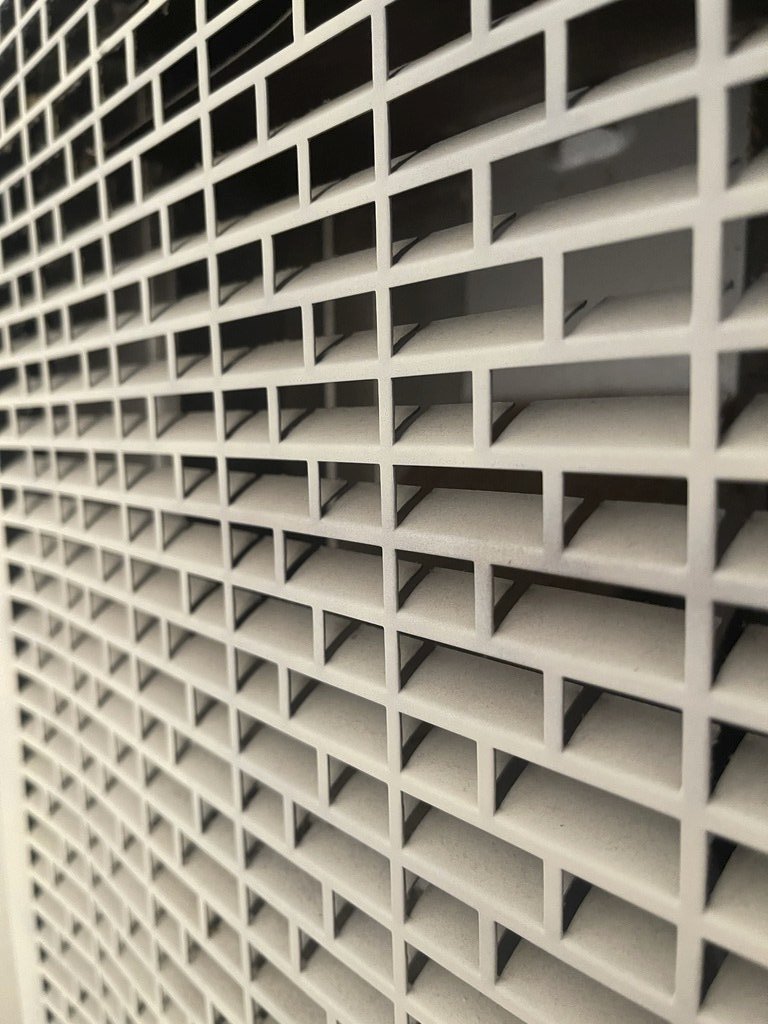}
};
\node (con) [left of=mid, yshift=-2cm,xshift=-2cm]{
\includegraphics[width=.2\textwidth,height=.2\textwidth]{pics/dandelionpic.jpg}
};
\node (out)[right of=mid,xshift=5cm]{
\includegraphics[width=.4\textwidth]{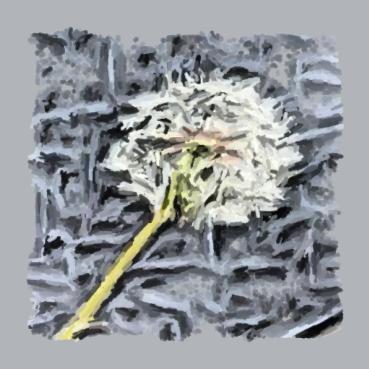}
};
\draw [arrow] (ins) -- (mid);
\draw [arrow] (con) -- (mid);
\draw [arrow] (mid) --node[anchor=center,above,align=left]{baseline\\technique} (out);

\node[above of=ins,yshift=.6cm]{
Inspiration
};
\node[above of=con,yshift=.6cm]{
Subject
};
\node[above of=out,yshift=2.1cm]{
painting
};
\node[below of=mid,yshift=.0cm]{
imagination
};
\end{tikzpicture}
  \caption{Example of the inspiration-imagination procedure: An imagination image is generated by applying style transfer from the inspiration image onto the subject image. Subsequently, the baseline technique is employed to create an artwork based on the imagination image. In this example the coloring from the original subject was mapped onto the imagination image by using the hue and saturation of the subject image in HSV space.}
  \label{inspillustration}
\end{figure}

\section{Inspiration-driven procedure}

We propose a procedure to paint a \emph{subject} image using another natural image as \emph{inspiration}. In this procedure we apply the style transfer of \cite{gatys2016image} from the inspiration image onto the subject image to create a new image which we call \emph{imagination}. We then apply a \emph{baseline technique} to create an artwork based on the imagination image.

In \cref{firstpic} we have taken the baseline technique to be the model trained through our medium-perception-driven procedure described above. This way the art style is created without being trained on human-made art.

%
%
%
%



\section{Technical details}

\subsection*{Convolutional brush engine}
To implement our medium-driven procedure we implement a paintbrush engine whose input, the \emph{action} is represented by a set of 3-tensors $a\in\mathbb R^{n\times n\times d}$ with the first two dimensions being the spatial dimensions of the image. This allows us to represent the artist $A_\theta$ as a convolutional neural network.

The main features in our representation of a brush stroke are:
\begin{enumerate}
    \item A \emph{direction field} which associates a $2\times 2$ projection matrix $P_\x$ to each (discretized) planar coordinate $\x=(x,y)$. The $\lambda=1$ eigenspace of the projection $P_\x$ represents the direction of a brush stroke through $\x$, if one exists. We use this projection-valued direction field instead of a vector field because we wish to let the distinction between the forward/backward directions be decided at the start of the brush stroke.
    To generate $P_\x$ in a way that respects this symmetry of the direction field we generate a $n\times n\times 2\times 2$ tensor of symmetric matrices $(A_\x)_{ij}$ and define $P_\x$ by shifting the spectrum of $P_\x=(A_\x-\lambda_0(A_\x)I)/(\lambda_1(A_\x)-\lambda_0(A_\x))$.

    \item A sequence of starting coordinates and starting directions.
\end{enumerate}

To generate a starting position with a convolutional network we let the output of the artist $A_\theta$ include a scalar field. We take the softmax of this field to obtain a probability distribution $\pi$ and obtain the starting position by sampling from $\pi$.

\subsection*{Differentiability}
There are several points in the construction of the brush engine where a na\"ive approach would render the medium non-differentiable with respect to the action $a$. We now describe these points and how we circumvent them. The \emph{straight-through} operation \cite{bengio_estimating_2013,yin_understanding_2019,van2017neural} is defined by applying a function $f$ in the forward pass but skipping it in the gradient computation. Define the \emph{two-input} straight-through operation as 
\[\operatorname{straight-thru}(x,y)=x-\operatorname{stop-grad}(x)+\operatorname{stop-grad}(y).\]
That is, the forward pass of $z=\operatorname{straight-thru}(x,y)$ is computed as if $z=y$ while the back-propagation is computed as if $z=x$. The standard definition of the stop-gradient corresponds to letting $y=f(x)$ for some function. Let $h$ be a scalar field representing the pixel values of the brush stroke. We replace $h$ with $\operatorname{straight-thru}(f,h)$ where $f$ is a corresponding brush stroke with soft edges. More interestingly, to make the probability distribution $\pi$ trainable we replace the brush stroke $h$ with
\[\operatorname{straight-thru}(\pi(\x_0)*h,h),\]
where $\pi(\x_0)$ is the probability of the sampled starting point. This allows the loss gradient for the brush stroke to propagate back through the probability distribution for the starting point.

To trace out a brush stroke $\x_0,\ldots,\x_k$ starting at $\x_0$ we iteratively read the direction field $P_{\x_i}$ at the current position $\x_i$ to obtain the next direction $v_{i+1}$. To do this we transform $\x_i$ into a one-hot representation and take the overlap with the direction field $P$ along the spatial dimension. It is important to compute $\x_i$ as $\x_i=\x_0+v_1+\ldots+v_{i}$ and not using the one-hot representation of $\x_{i-1}$ in order to let the gradient propagate through all the the directions of the brush stroke.

\begin{figure}[h]
  \centering
  \begin{tikzpicture}
\node (artist) [wideblock1]{artist+medium $M\circ \tilde A_\theta$};
\node (content) [wideblock2,left of=artist, yshift=2cm,xshift=-4cm]{
subject
};
\node (a1) [wideblock2,left of=artist,xshift=-4cm]{
canvas
};
\node (a2)[wideblock2,right of=artist,xshift=4cm]{
canvas'
};
\draw [arrow] (content) -- (artist);
\draw [arrow] (a1) -- (artist);
\draw [arrow] (artist) -- (a2);
\end{tikzpicture}
  \caption{We model the artist as a map which is applied iteratively. A single iteration is hown here.}
  \label{iterative}
\end{figure}
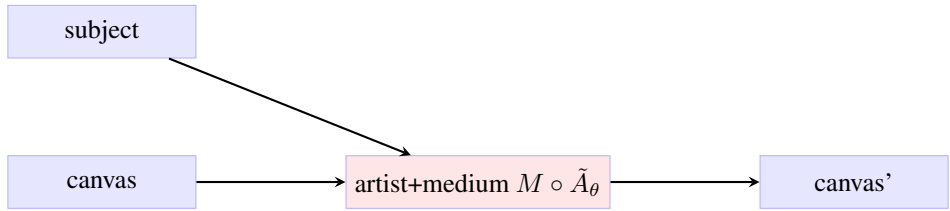

\subsection*{Iterative artist}
\begin{figure}[h]
  \centering
\includegraphics[width=.4\textwidth,height=.4\textwidth]{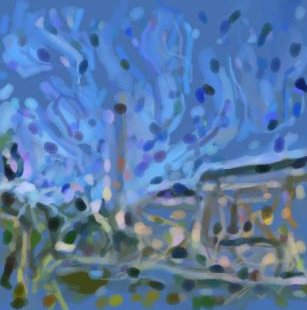}
\includegraphics[width=.4\textwidth,height=.4\textwidth]{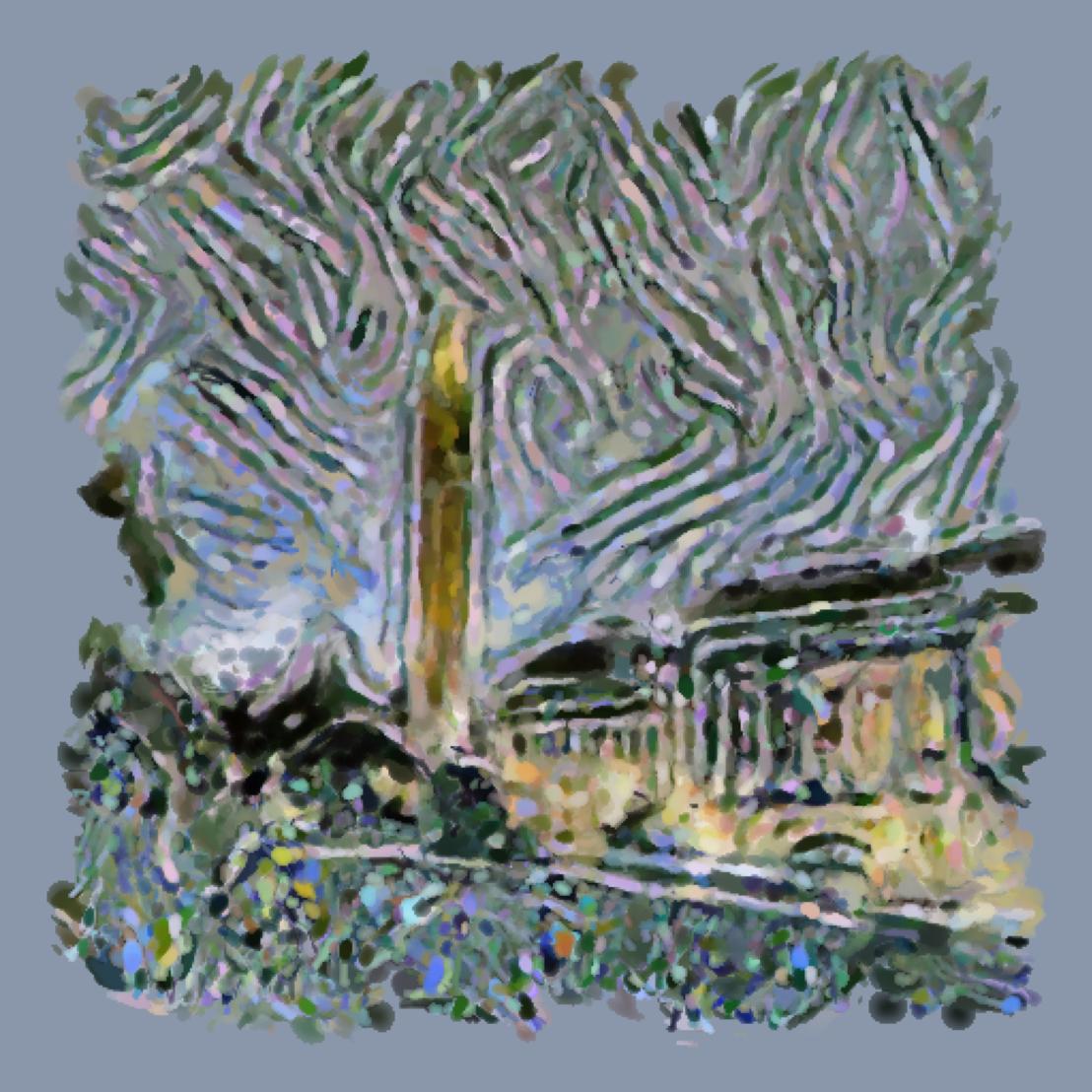}
  \caption{Two paintings made by combining our two methods, using the artist from the medium+perception-driven procedure as the baseline technique for the inspiration procedure. The painting on the right was made with several iterations of the iterative artist working on smaller patches of the image, resulting in the difference in style.}
  \label{patchwise}
\end{figure}

In order to facilitate training we model the artist as a parameterized map $\tilde A_\theta(c,s)$ that takes two inputs: the subject and a canvas with the artist's own unfinished painting \cref{iterative}. $M\circ\tilde A_\theta$ is iteratively applied, beginning from a blank canvas, in order to create the painting. The artist also chooses the color of the blank background. The fact that the background color is not fixed helps the artist learn to take the brackground into account when determining the next action. The trained iterative artist can be applied for any number of iterations and at different length scales, resulting in variations on the style (\cref{patchwise}).

\subsection*{Directional loss}
\label{dirloss}
In our medium-perception-driven procedure we optionally apply a direction loss between the subject image and the reconstruction, which we define in this section. 

Given a scalar-valued function $f$ on the plane, let $\nabla f$ be its gradient and let $\Gamma f=(-\partial_y f,\partial_x f)$ be the rotation of $\nabla f$ by 90 degrees. We view $\Gamma f$ as a row vector. Given a function $f$ with (color) channels $f_c$, define a matrix-valued function $\rho$ as the sum of outer products:
\[\rho(\x)=\sum_{\text{channel }c}\Gamma f_c(\x)^T\Gamma f_c(\x).\]
Let $S$ be a smoothing kernel. We then define the \emph{direction field} of the image described by $f$ as the 2x2-matrix-valued function $\tilde\rho=S*\rho$. 

We use 2x2 matrices to represent the direction field in order to gain sign-symmetry of the directions. This is important in the case of ripple-like $f$ (for example $f(x,y)=\cos(Ax+By)$) as nearby directions would otherwise cancel each other out.

We then define the direction loss between two functions using a scale-invariant loss:
\[\operatorname{Dirdist}(f_1,f_2)=\sqrt{1-\frac{\tilde\rho(f_1)\cdot\tilde\rho(f_2)}{\|\tilde\rho(f_1)\|\|\tilde\rho(f_2)\|}},\]
where the dot product and norms are entrywise (also known as Hilbert-Schmidt or Frobenius) and averaged over $\x$.
In practice $f$ is given as a finite $n\times n\times3$ tensor and we compute $\nabla f$ and $\Gamma f$ using finite differences. 

\textbf{Further information.} 
Code will be available at 
\url{https://github.com/nilin/art_ab_initio}.

%

%


\section{Discussion}

We have shown two ways in which an AI model can create art styles without using human art in the training data. It is thus possible to construct a generative model that has never seen a human artwork \emph{or} any outputs from other generative AI models. This is important because artworks generated with other AI models can and do leak human-made art styles from their training data \cite{carlini_extracting_2023}.



In this paper we have given a proof of concept to show that painting styles can be created which are not present in the training data. We do this by testing a simplified proxy for ``aesthetics'' which uses the inductive bias from the artistic medium with a reconstruction loss to allow for abstraction. We further proposed a way to direct the evolution of painting styles through inspiration from natural images which allows the generated painting styles to evolve without using artistic inputs as training data. We do not claim to capture or compete with the perception and sense of aesthetics of a human artist, which are highly complex (see \cite{gombrich_art_2000} for discussions which are beyond the scope of this paper). But we believe that our contribution is significant in that it objectively tests the widely held assumption that an AI is limited to interpolating human-made creations. 

We now speculate about possible implications of this work. As a proof of concept we have excluded artistic works from the training data in this paper, but in practice we envision that human creativity can be re-introduced. For example an artist could use a generative model for their own styles in place of the the baseline technique. In this way our method could be used as a creative tool to allow an artist to experiment with variations of their own styles.  

For users of generative art models our methods provide a way to ethically use such AI tools, ensuring in an objective way that they are not infringing on artists' copyright.
We hope that such a development could also be beneficial to artists and their ability to publicly share their work. Specifically we hope that the concern of having one's personal style reproduced by AI lessens if generative AI models become less reliant on artistic training data.

\section{Acknowledgements}
N.A. thanks Alina Scotti for literature references about aesthetics. 
N.A. was supported by the Simons Foundation under grant no. 825053. 

\bibliographystyle{plain}
\bibliography{library}

\end{document}